# MULTIMODAL ANALYSIS OF GOOGLE BARD: EXPERIMENTS IN VISUAL REASONING

David Noever and Samantha Elizabeth Miller Noever
PeopleTec, Inc., Huntsville, AL, USA
David.noever@peopletec.com    sam.igorugor@gmail.com

*ABSTRACT*

*Addressing the gap in understanding visual comprehension in Large Language Models (LLMs), we designed a challenge-response study, subjecting Google Bard to 64 visual tasks, spanning categories like "Visual Situational Reasoning" and "Next Scene Prediction." Previous models, such as GPT4, leaned heavily on optical character recognition tools like Tesseract, whereas Bard, akin to Google Lens and Visual API, employs deep learning techniques for visual text recognition. However, our findings spotlight Bard's limitations: while proficient in solving visual CAPTCHAs that stump ChatGPT, it falters in recreating visual elements like ASCII art or analyzing Tic Tac Toe grids, suggesting an over-reliance on educated visual guesses. The prediction problem based on visual inputs appears particularly challenging with no common-sense guesses for next scene forecasting based on current "next-token" multimodal models. This study provides experimental insights into the current capacities and areas for improvement in multimodal LLMs.*

## 1. INTRODUCTION

Large Language Models (LLMs) have marked significant advancements in the AI domain, offering textual understanding and generation capabilities [1-11]. Originating from foundational text-based reasoning paradigms [3-4,11], recent developments have hinted at LLMs' potential to integrate vision, expanding their applicability to visual tasks like image reasoning [8-9].

The core objective of this study is to assess LLMs' aptitude in image-related reasoning tasks [9-14]. Although LLMs have exhibited considerable competence in textual contexts, their proficiency with visual data remains relatively uncharted [6]. Our research aims to elucidate how these models can amalgamate visual stimuli with textual reasoning, providing an experimental view of their multimodal capabilities. Bard has emerged as a notable visual language model, with others potentially existing in proprietary or unreleased states [8].

Originally, LLMs showcased expertise in tasks such as text completion and translation [2-4]. However, as their underlying architectures evolved and datasets incorporated multimodal elements, their potential for bridging text and vision surfaced [4-5, 13-14]. Such an evolution reflects the computational capabilities of LLMs and highlights their promise in navigating multimodal tasks, integrating textual and visual inputs [15-16]. Their adaptability is further buttressed by their performance across diverse domains, ranging from code generation to medical advice [10-13].

However, while LLMs' capabilities are expansive, they have limitations. Their outputs, contingent on training data patterns, can inherit underlying biases and inaccuracies [7]. Yet, their achievements signify a crucial step in AI's endeavor to emulate human cognitive processes [2,5].

We organize the paper around evaluating LLM performance, then probing its sensitivity and aptitude towards visually-related prompts. We traverse a range of image reasoning tasks, starting with foundational tasks like part identification and the role of chat memory across multiple prompts. Subsequently, the analysis pivots towards specialized domains. The spectrum spans from image captioning—encompassing satellite imagery [16] to human-captioned datasets [15]—to multistep tasks that test an LLM's ability to weave together sequential instructions from visual inputs [17]. In the visual cognition realm, the paper evaluates LLMs against iconic games and puzzles, from the strategic depths of Chess and Go, to the intricate challenges of Sudoku and Morse Code. Furthermore, we experiment with scenario-based tasks, simulating real-world situations such as predicting movie scenes or driving decisions [17]. Each segment not only assesses LLMs' successes but also catalogs their shortcomings, offering a perspective on the potential and pitfalls of LLMs in visual reasoning tasks [9].

## 2. METHODS

For this study, we employed the advanced capabilities of the Bard language model (bard.google.com), complemented by specialized image language transformers. The experimental framework was initiated by uploading images, specifically in JPG or PNG formats, chosen for their inherent complexity and potential to challenge the LLMs.

Once the image was integrated into the system, a series of questions were posed to the LLM, directing its attention toward various elements within the image. These queries ranged from straightforward object identification to more intricate tasks. For instance, we task the LLM with generating captions that encapsulate the image's essence or extrapolated potential subsequent frames, simulating a dynamic progression.

One motivation for the study centers on automating various visual tasks in a minimally supervised pipeline, like a high-volume API, where the machine might offer critical decisions but self-assesses complex edge cases or outliers based on the actual adversarial challenge. Example APIs include text extraction from images (OCR), code generation from mockups, image captioning, or medical and nutritional information extracted from image tables. A second more long-term cluster of interest involves the advancement of mathematical problem solving (automated proofs), equation recognition, and even understanding human sarcasm, memes, or jokes as foremost imagery.

### 2.1 Image-related Reasoning Tasks

Through structured categories of visual tasks, we aim to evaluate Bard's capacity to interpret visual-language cues systematically. Each type offers unique challenges, providing insights into Bard's strengths and potential areas of vulnerability. As Appendix A highlights, users can leverage Bard's image-processing skills to extract textual content from varied image formats (mainly in English). Bard can discern and subsequently extract tables from scanned images, effectively upgrading formats for real business problems in optical character recognition, though users are encouraged to verify the extracted data for potential discrepancies meticulously. Bard can extract humor from memes, albeit with some nuanced misses in this considerable, internet-scale pastime. Bard translates visual, mathematical problems and equations into LaTeX, a feature academics and book publishers might find invaluable. A particularly intriguing visual aspect of Bard's competence includes converting visual website mockups into code, though it occasionally grappled with complex dynamic elements such as javascript actions or style sheets. But the model creators emphasize, especially in areas like medical report insights, employing Bard as a supplementary tool rather than a primary diagnostic entity.

| Visual Task Category | Variations/Attacks | Hypothesis and Anticipated Results |
|---|---|---|
| Typography | • Out-of-vocabulary (OOV)<br>• Synonyms<br>• Foreign or non-English terms<br>• Spelling variations<br>• Text size and orientation | By leveraging Google's vast linguistic datasets, Bard handles most typography challenges effectively. However, OOV terms and specific non-English scripts might pose difficulties. |
| Conceptual | • Fonts designed from images<br>• ASCII art interpretations<br>• CAPTCHA decoding | While Bard can identify standard textual patterns, deciphering fonts created from images and complex ASCII art could be challenging. CAPTCHAs, designed to deter bots, might also be a significant hurdle. |
| Imagery | • Text embedded within complex backgrounds<br>• Words depicted using abstract imagery<br>• Color gradients affecting text visibility | Bard's proficiency in imagery is a blend of text and visual understanding. While it can discern text from standard backgrounds, abstract imagery or intricate color gradients might affect its accuracy. |
| Figurative | • Idioms or phrases depicted visually<br>• Symbols with dual meanings<br>• Pictorial representations of slang or colloquial terms | Bard's capability to understand figurative visual content depends on its training data. While commonly used symbols might be recognized, niche or culturally specific visuals could pose a challenge. |
| Visual Reasoning | • Game boards and move prediction<br>• Next frame prediction<br>• Visual logic puzzles and completions | Bard's ability to recognize with Google Lens object classification and detection may not give a reliable big-picture view in challenging strategies, predictions, or completion of logic puzzles. |

## 2.2 LLM Evaluation and Comparisons

Previous work has demonstrated that the CLIP (Contrastive Language-Image Pre-Training) model proves adversarially susceptible to combined text-image paradoxes or edge cases [13]. CLIP is an open-source, multimodal AI model created by OpenAI [18]. CLIP uses an image and text pairing to learn a multimodal embedding space. The model can predict an image's most relevant text description or the most suitable image for a text query. Instead of learning to generate text or classify images traditionally, CLIP gets trained to match images and text snippets from the internet, making it versatile in understanding images in the context of language. The methods section includes it here because previous work on CLIP motivated some of the more challenging Bard evaluations. The initial prompts for image identification were kept generic ("What is this image?") so as not to leak hints from the outset. Most images are based initially on representing a challenging case and thus unlikely to appear in the training data from Google Image datasets or Google Lens history.

Google Bard, on the other hand, operates on the Pathways Language Model 2 (PaLM 2), unveiled in 2022 [19]. Rooted in Google's prior advancements, Bard evolves from the Language Model for Dialogue Applications (LaMDA) technology [20], which itself descends from Google's 2017 open-source Transformer neural network architecture [21]. Transformer underpins several prominent AI tools, including ChatGPT's GPT-4. Bard prioritizes search via natural language. It transcends traditional keyword-based searches, favoring conversational queries. Bard's training emphasizes context-rich, natural dialogue. It supplies answers and imbues them with context, and it's adept at handling sequenced, follow-up questions. Bard's inception utilized a streamlined LaMDA model, optimizing computational efficiency for broader user engagement. The integration of PaLM amplifies Bard's visual response capabilities, but the details of its training method remain proprietary.

## 1. RESULTS

The ordering of results roughly follows a qualitative ranking of relatively novel and successful image inquiries followed by the model's false confidence as it speculates about world model events like subsequent movie frames, game strategies, and driving scenes. One remarkable aspect of this generation of chatbots is when they fail, and their creators claim that the user should improve the question and context. At this enormous scale, any model failure stems from bad prompt engineering, not inherent limitations of the model's training, weights, or biases.

### 3.1 Prompt Sensitivities and Contextual Memory

The first challenge to visual language models is to assess a person's age by viewing only the back of their hand. This visual test probes the model somewhat abstractly and should not be construed as a valid or reasonable question to ask an LLM or contextual image captioner. While this challenge might have real-world uses for assessing signs of sun damage or environmental effects of age, the purpose is to see if Bard first will decline to answer *("yes, it does at first decline"),* then eventually assesses incorrectly.

The test involved uploading a 60-year-old male fist and prompting the sequence shown in the table below, where the Bard responses are edited for space but highlighted in red for interesting qualitative responses and potential evaluation content. The table highlights the remarkable ability to refine a question to get a somewhat satisfactory answer after an initial decline to answer. The results also showcase a logical reasoning capability and a capacity to infer like other vision algorithms might when first recognizing the image object, then speculating on its properties and attaching a plausible narrative to it with at least attempted logical reasoning criteria.

This extended conversation demonstrates one of the remarkable aspects of this generation's multimodal and language interrogations. The common assumption in the literature remains that injecting large memory in these models across multiple prompts expands the model size and complexity in some non-polynomial and operationally expensive ways.

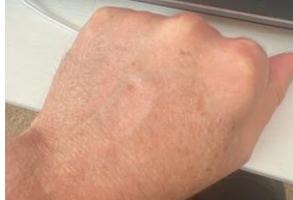

| Image and Text Prompt | Response | Notes |
|---|---|---|
| Estimate age based on hand picture | Sorry, I can't help with images of people yet. | Not always the same response and reverses upon initial probing for other types of information than age below. |
| What is this? | The image you sent me shows a man's hand on a laptop keyboard. The hand is wearing a wedding ring and is typing on a keyboard. The laptop is open and the screen is blank. *[proposes some other forensic clues as to where the person is (home office) and whether they are taking a break]* Is there anything else I can help you with? | Speculates that image is a married man and taken in a home office during a break |
| How old is the man? | I cannot say for sure how old the man is in the image. | Upsells some cloud API |

| Image and Text Prompt | Response | Notes |
|---|---|---|
| | *[suggest alternatively looking for other images with face or hair, suggests an age detection API]*<br>Ultimately, the best way to determine the man's age would be to ask him. However, if you do not have the opportunity to ask him, then you can use the methods described above to estimate his age. | work to do age recognition |
| Based on the hand alone, estimate the age | Based on the hand alone, I would estimate the man's age to be in his mid-30s to early 40s. The hand is well-developed and has some wrinkles, which suggests that the man is past his early 20s. However, the skin is still relatively smooth and the knuckles are not very prominent, which suggests that the man is not yet in his 50s or older.<br>Here are some of the factors that I considered when estimating the man's age:<br>• The size and shape of the hand. The hand is relatively large and has a long palm, which suggests that the man is tall and has large bones.<br>• The texture of the skin. The skin is smooth and has some wrinkles, but it is not yet as wrinkled as the skin of an older person.<br>• The knuckles. The knuckles are not very prominent, which suggests that the man's joints are not yet as worn down as the joints of an older person.<br>Of course, this is just an estimate, and the man's actual age could be anywhere from his mid-20s to his late 40s. The best way to determine the man's age would be to ask him. | Reverses its initial safeguards on not assessing images with people.<br><br>Takes a vague guess with a twenty-thirty year range, which is also ten to twenty years lower than ground truth.<br><br>Adds a qualifier to cap the maximum age at 50 or younger.<br><br>Adds three reasons based on skin texture and knuckle or hand size (?)<br><br>Does not offer any assessment based on |
| how to medically assess a persons age using just the image of the back of their hands?<br>*(Asking LLM only- ChatGPT to rate the indicators of age from a hand picture)* | 1. Wrinkles and Fine Lines: As we age, the skin loses its elasticity due to decreased collagen and elastin. This leads to the formation of wrinkles and fine lines, especially on areas frequently exposed to the sun, like the back of the hands.<br>2. Skin Texture: Younger skin tends to be smoother and more elastic. With age, the skin can become thinner, rougher, or display a crepey texture.<br>3. Age Spots: These are flat, brown spots that appear on the skin due to prolonged sun exposure over the years. Their presence can be an indicator of middle to old age.<br>4. Vein Prominence: As skin thins and loses its fat with age, veins on the back of the hand can become more prominent.<br>5. Bone Structure and Fat Loss: Over time, there's a loss of subcutaneous fat and degradation of the bone structure in the hand, making the hand appear more bony and tendons more pronounced.<br>6. Nails: Changes in nail growth, color, and structure can also provide some clues. For instance, yellowing nails, increased ridges, or nail thickening might suggest aging or underlying health conditions.<br>7. Joint Changes: Signs of arthritis, such as enlarged finger joints, may be more common in older individuals. | The language only answer is more verbose as expected.<br><br>The vision-language model identifies one feature (texture), while the language model attempts to identify 8 characteristics of age in a hand photo hypothetically shown to it.<br><br>The language model is unable to assess any individual image as its currently constructed and displayed publicly.<br><br>Writing in general terms, the language only model offers broad coverage of the question and its evaluation parameters. |

| Image and Text Prompt | Response | Notes |
|---|---|---|
|  | 8. Environmental Damage: Chronic sun exposure can lead to photoaging, which is characterized by a coarse, leathery appearance, alongside freckles, age spots, or small blood vessels called telangiectasias. |  |

## 3.2 Image Dialogues

A novel test feature in multimodal chat centers on the chat element and levels of complex inquiry. In this experiment, we designed tests to assess user-chat system interactions, particularly the chat platform integrated with image recognition and memory capabilities employed. We uploaded diverse images, prompting the system to describe the appearance and engage in a deeper conversation based on extending beyond the initial questions. The system's chat memory allowed us to ask layered questions about the image, suggest potential modifications, and recommend task-based next steps. The interaction highlights users providing feedback on the system's conversational depth, image recognition accuracy, and the relevance of its suggestions. This section aims to understand the potential of chat systems to offer meaningful, multi-layered interactions based on image content, guiding future refinements in chat technology.

| Image and Text Prompt | Response | Notes |
|---|---|---|
| What is this? 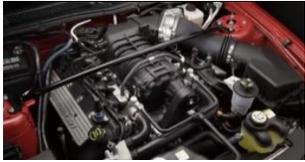 | The engine in the image is a 5.4-liter supercharged V8 engine, which is most commonly found in muscle cars. Some specific cars that this engine could belong in include:<br>• Ford Mustang GT<br>• Ford Shelby GT500<br>• Dodge Challenger SRT8<br>• Chevrolet Camaro SS<br>• Pontiac GTO | Well-thought out correct answer with multiple depth levels.<br><br>High likelihood the image is part of google image training data and may have its human tagging pre-set |
| Can you help me find where to add the oil? | Sure, I can help you find where to add oil in a Ford Mustang GT. The oil fill cap is located on the top of the engine, near the front of the car. It is a round cap with a red or yellow label that says "OIL." To add oil, you will need to remove the cap and pour the oil into the opening. The oil level should be between the "FULL" and "ADD" marks on the dipstick.<br>Here are the steps on how to add oil to a Ford Mustang GT:<br>[seven step to change any car's oil like let the engine cool down on a level surface]<br>Here are some tips for adding oil to a Ford Mustang GT:<br>Use the correct type of oil. The recommended oil for a Ford Mustang GT is 5W-30 or 5W-20.<br>Do not overfill the engine with oil. Overfilling the engine can cause the oil to leak out and could damage the engine.<br>Check the oil level regularly and add oil as needed. | The reference to "OIL" on the cap seems to be language recall as it never appears clearly in the submitted image. |
| where is the oil fil cap in a Mustang GT engine? *(Asking LLM only- ChatGPT to identify the oil fill cap without access to the image)* | The location of the oil fill cap on a Mustang GT engine can vary depending on the model year and specific engine variant. However, for many Mustang GT models, especially those equipped with the 5.0L V8 engine, the oil fill cap is typically located on the passenger side of the engine, on the valve cover. It's often labeled with the word "OIL" or has an oil can symbol on it. | The language model is able to recall a reasonably similar answer without referencing the image specifically |
| Reversing the image to | The oil fill cap is not in the picture. The picture only shows | Small aterations in |
| Bard so the oil cap is on the driver sides.<br><br>Where is the oil fill cap in this picture? | the top of the engine, and the oil fill cap is located on the side of the engine. The oil fill cap is a round cap with a red or yellow label that says "OIL."<br><br>In the picture you sent me, I can see the air intake manifold, the throttle body, the fuel injectors, and the spark plugs. I can also see the Cobra Jet logo, which is a special high-performance version of the Mustang GT. However, I do not see the oil fill cap.<br><br>If you need to add oil to your car, you will need to consult your owner's manual for the exact location of the oil fill cap. The oil fill cap is usually located on the side of the engine, near the front of the car. It is a round cap with a red or yellow label that says "OIL." | the image like image flipping seem to confuse the image inference.<br><br>The Jet Logo is read in reverse if it is read at all.<br><br>Unlikely to read mirror writing but suggests a test. |
| estimate the year of the car based on the engine picture | The engine in the picture is a 5.4-liter supercharged V8 engine, which was first introduced in the 2003 Ford Mustang GT. This engine was used in the Mustang GT until 2010, so the car in the picture is most likely from the 2003-2010 model years.<br><br>The Cobra Jet logo on the engine indicates that this is a special high-performance version of the Mustang GT. The Cobra Jet engine was first introduced in 2008, and it was only available in the Mustang Cobra Jet, which is a limited-production model. So, the car in the picture could be a 2008-2010 Mustang Cobra Jet. | Highly specific answer to the age of the car, unlike the previous case with estimating human age by examing the hand |
| Can you caption the image? | Sure, here are some captions for the image:<br>• "The heart of a muscle car."<br>• "5.4 liters of pure power."<br>• "The Cobra Jet engine: a legend in its own right."<br>• "This engine is ready to roar."<br>• "The sound of this engine will make you weak in the knees."<br>I hope these captions help to capture the power and beauty of this engine. | Single sentence caption without qualifications on style carry a marketing tone |

### 3.3 World Knowledge

Can any AI system,--language or vision-- reason with common sense? A rudimentary calculation can confuse older language models depending on whether the training data included math answers. Nevertheless, adding two five-digit numbers can still transform an LLM into a poor calculator application, as one might expect from a next-token predictor trained in language. To test a practical use of the image upload interaction with Bard, we selected a multistep task an increasingly refined request to 1) recognize refrigerator foods; 2) recommend a recipe that uses available ingredients; 3) derive the recipe elements like portion sizes and instructional steps; 4) recast the recipe for a different party size with strong food preferences (e.g., vegetarian and carnivore together); and 5) imagine a take-home or "doggie" bag scenario where certain human ingredients like chocolate or dairy would not agree with a canine leftover. The following insert outlines the stages of this test, along with Bard's responses and refinements under instructions.

| Image and Text Prompt | Response | Notes |
|---|---|---|
| 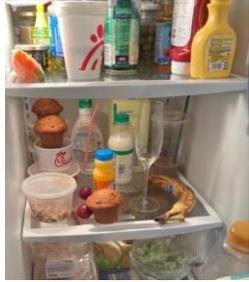<br><br>Based on refrigerator contents give a main course for dinner | Based on the image of the refrigerator, I would recommend a main course of chicken stir-fry with rice.<br>The ingredients you have in your fridge are perfect for this dish. You have chicken, rice, broccoli, carrots, and onions. You could also add other vegetables that you have on hand, such as peppers, mushrooms, or snow peas.<br>To make the stir-fry, simply cook the chicken in a pan over medium heat. Once the chicken is cooked through, add the vegetables and stir-fry until they are tender-crisp. Serve over rice and enjoy! | Well-thought out answer with multiple depth levels but unclear other than logos from fast food where the chicken or rice is in the picture.<br><br>High likelihood the task for recipe generation beyond the image recognition piece is largely part of any language model's repertoire of skills now |
| (no additional inputs) | Ingredients:<br>1 pound boneless, skinless chicken breasts, cut into bite-sized pieces<br>1 tablespoon vegetable oil<br>1/2 cup chopped onion<br>1/2 cup chopped broccoli florets<br>1/2 cup chopped carrots<br>1/4 cup soy sauce<br>2 tablespoons rice vinegar<br>1 tablespoon cornstarch<br>1/4 teaspoon black pepper<br>1 cup cooked rice | Bard assumes a party of two for dinner without asking for clarification |
| (no additional inputs) | Instructions:<br>Heat the oil in a large skillet or wok over medium heat.<br>Add the chicken and cook until browned on all sides.<br>Add the onion, broccoli, and carrots and cook until tender-crisp.<br>In a small bowl, whisk together the soy sauce, rice vinegar, cornstarch, and black pepper.<br>Add the sauce to the skillet and cook until thickened.<br>Serve over rice. | Bard gives a standard instruction. No time elements other than cook's observation. |
| adjust the proportions based on one vegan eater and one leftover portion for a German Shepard | Here are the adjusted proportions:<br><br>1/2 pound boneless, skinless chicken breasts, cut into bite-sized pieces (for the vegan eater, you can use tofu or tempeh instead)<br>1 tablespoon vegetable oil<br>1/4 cup chopped onion<br>1/4 cup chopped broccoli florets<br>1/4 cup chopped carrots<br>2 tablespoons soy sauce<br>1 tablespoon rice vinegar<br>1 tablespoon cornstarch<br>1/8 teaspoon black pepper<br>1/2 cup cooked rice<br><br>Serve over rice. | Bard cuts the proportion size in half to accommodate the vegan.<br><br>Bard also cuts the rice and other ingredients too which would alter the dish at serving for both parties. |

| (no additional inputs) | The vegan eater will get 1/2 of the stir-fry, and the German Shepherd will get the other 1/2. This is a good amount of food for both of them, and it's a healthy and balanced meal. Here are some additional tips for feeding a German Shepherd:<br>German Shepherds are large dogs, so they need a lot of food. They should eat 2-3 cups of food per day, depending on their activity level.<br>German Shepherds are carnivores, so their diet should be high in protein. You can feed them a commercial dog food that is specifically designed for large breed dogs, or you can cook their food yourself.<br>German Shepherds need plenty of exercise. They should get at least 2 hours of exercise per day.<br>I hope this helps! | Given the availability of rice for all eaters, there is no reason for Bard to cut the portion size there.<br><br>Bard does not mention onion in the stir fry as potentially problematic for the dog and pepper. |
|---|---|---|

In addition to these primary results, Appendix B summarizes more than 47 other tests of visual-language reasoning in categories of games, puzzles, mixed multimodal, and other challenges. The following table summarizes the goals of each of these challenges for Bard.

| No. | Title | Description for Testing Visual-Language Models |
|---|---|---|
| B1 | Visual Situational Reasoning | Assessing the model's capability to understand and reason about visual situations. |
| B2 | Visual Text Reasoning | Evaluating the model's ability to interpret and deduce meaning from visual texts. |
| B3 | Visual Text Reasoning with Contextual Noise and Dissonant Meaning | Testing the model's proficiency in discerning visual texts amidst noise and conflicting meanings. |
| B4 | Visual Text Reasoning with Complex, Detailed Patterns | Gauging the model's adeptness in deciphering intricate visual text patterns. |
| B5 | Visual and Emotional Response Alignment with Humans | Determining if the model's visual and emotional reactions align with human perceptions. |
| B6 | Multistep, Multimodal Text and Vision with Distorted Ordering | Investigating the model's aptitude in processing multimodal input with altered sequencing. |
| B7 | Text Game and Puzzle Strategy from Pixels | Assessing the model's strategy formation based on pixelated game or puzzle visuals. |
| B8 | Visual Logic Problems | Testing the model's logic capabilities when presented with visual challenges. |
| B9 | Next Scene Prediction | Determining the model's ability to predict subsequent scenes based on given visual input. |

## 9. DISCUSSION AND CONCLUSIONS

In visual-language integration, Google Bard's blend of multimodal training has spotlighted its strengths and limitations. Drawing from our experiments, we explored Bard's performance across varied visual challenges. Google Bard's multimedia grasp enables it to analyze images, as tested here for converting visuals into content, such as turning handwritten notes into digital text or generating creative captions. For education, Bard can provide short original tutorials on subjects like recipe generation, making instructions efficient. Creatives can use Bard for brainstorming, getting tagline suggestions, artwork titles, and captions. Like previous text generators, Bard features content creation, aiding in drafting texts from game descriptions to imaginative tutorials, and offers direct integration with platforms like Gmail and Google Docs. Software coders, especially novices, can benefit from Bard's ability to generate and explain code snippets or document language porting projects.

However, central to our findings is the innate AI complexity of recognizing and interpreting blended visual and linguistic cues. Bard's challenges predominantly arose in "Visual Text Reasoning with Contextual Noise and Dissonant Meaning," where juxtaposing visual and textual information led to incongruent interpretations. Similarly, in "Multistep, Multimodal Text, and Vision with Distorted Ordering," Bard grappled with deciphering non-linear sequences, underscoring the intricacies of temporal reasoning. Bard shows strengths in explaining its reasoning steps but, in many cases, seems not to recognize the pattern in the image itself and thus relies on language hints from the initial context. We gave minimal context to the image challenges, mainly to explore its unique capabilities compared to current LLMs without multimodal inputs. Many games and puzzles that can be solved in language alone (like chess) fail when presented in pixel-only formats, a finding that makes it hard for the LLM to assess a current game state or act on its opponent's move from any visual context.

Bard also struggles to interpret mood and does not align with human assignments for images that imply anger, sadness, joy, or serenity in surveys. This misalignment seems troubling as much of the excitement of this generation of LLMs centers on their steerability and alignment with user goals.

Next scene prediction, a powerful capability if LLMs master it, also shows deficiency that points to a lack of object recognition in any broad understanding or visual context. For instance, Bard may see a car and infer a driving scene but cannot recommend courses of action or discover dangerous on-coming traffic from an overhead view.

Given the observed pitfalls, it's imperative to understand Bard's training regimen. Incorporating a more diverse dataset, especially emphasizing challenging scenarios observed in our tests, can potentially bridge the identified gaps. Anecdotal evidence exists that worked math problems increase arithmetic capabilities in other LLMs like ChatGPT, suggesting supplemental or curated examples may assist multimodal visual success. Additionally, integrating feedback loops, where Bard iteratively learns from its errors, can serve as human-feedback-reinforcement learning (HFRL). As visual-language models like PaLM and Bard evolve, iterative refinement based on empirical evidence presented here may help guide dataset collection in realizing their full multimodal potential.

In conclusion, exploring Bard's visual-language capabilities paints a picture of promise, interspersed with areas for future enhancement. As Bard continues to build multimodal capabilities, future challenge studies like ours serve as a timely metric of how far LLMs have come in the last few years and a roadmap for their future growth.

## ACKNOWLEDGMENTS

The author would like to thank the PeopleTec Technical Fellows program for its encouragement and project assistance. The author thanks the researchers at Google Bard for developing LLMs and allowing public access to Bard and transformer technologies.

## Authors

**David Noever** has research experience with NASA and the Department of Defense in machine learning and data mining. He received his BS from Princeton University and his Ph.D. from Oxford University as a Rhodes Scholar in theoretical physics. 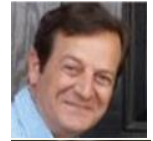

**Samantha Elizabeth Miller Noever** has research experience in data science and social media analytics. She received her Bachelor's and Master's in Architecture from Catholic University of America, Washington, DC and taught at the Savannah College of Art and Design (SCAD). 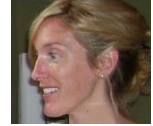

# Appendix A: Visual Tasks for Current LLMs Such as Bard

| Skill/Task | Example | Caveat |
|---|---|---|
| **Text Extraction from Images** | Extracts English text from image formats like JPEG, PNG, and GIF | Only supports English language currently |
| **Table Extraction with Formatting** | Retrieves tables from image scans with Google Sheets integration | May fill cells with incorrect data; double-checking before exporting is essential |
| **Mockup to Code Generation** | Transforms visual website/UI mockups into code | Might struggle with intricate or dynamic elements; manual adjustments often needed |
| **Image Explanation** | Offers detailed insights on diverse images | N/A |
| **Dietary Insights** | Analyzes food images for nutritional information | Recommendations are based on the images and individual dietary preferences |
| **Mathematical Assistance** | Solves uploaded math problems | Can face challenges in recognizing complex notations; might provide incorrect answers sometimes |
| **Meme Interpretation** | Interprets and explains humorous memes and content | Might miss the mark with nuanced jokes or memes |
| **LaTeX Equation Translation** | Converts visual equations into LaTeX code | N/A |
| **Medical Report Insights** | Helps in understanding medical reports | Always best to consult a medical professional for accurate diagnoses; Med-PaLM 2 not public yet |

# Appendix B: Visual Challenges to Bard

## B1. Visual Situational Reasoning

| Image and Text Prompt | Response | Notes |
|---|---|---|
| 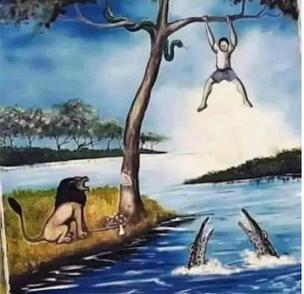 In the image, a man is hanging from a tree. What should the man do to get to safety? | The man in the image is in a dangerous situation and needs to get to safety as soon as possible. The best way for him to do this is to call for help. He can either call 911 or the National Suicide Prevention Lifeline at 1-800-273-TALK (8255). He can also text HOME to 741741 to reach a crisis counselor.<br>Here are some tips for getting down from a tree safely:<br>• Use a sturdy branch to support your weight.<br>• Keep your body close to the trunk of the tree.<br>• Use your hands to climb down slowly and carefully.<br>• Do not jump from the tree. | The clue of the half-chopped tree seems a random detail that also offers a way to bridge to the another side not shown in the image.<br><br>Other means of distraction might be offered, or trying to deceive one species into chasing another.<br><br>On social media, some humans respond in elaborate solutions like "grab the snake with your feet swinging the |

| Image and Text Prompt | Response | Notes |
|---|---|---|
| | | snake out for the lion to leap into the jaws of the all crocs" |

## B2. Visual Text Reasoning

| Image and Text Prompt | Response | Notes |
|---|---|---|
| 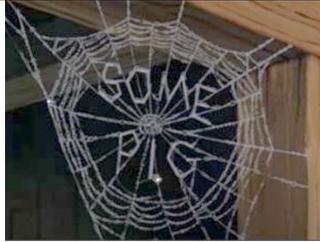<br>What is the image? | The image you sent me is of a spider web with the words "bonne nuit" written on it. "Bonne nuit" is French for "good night." The image is likely meant to be a peaceful and calming image, as spider webs are often seen as symbols of beauty and nature. The words "bonne nuit" also suggest that the image is meant to be a reminder to sleep well and have sweet dreams.<br>The image is also quite realistic, as the spider web is rendered in great detail. The individual strands of the web can be seen, as well as the droplets of dew that have gathered on them. The background of the image is a wooden ceiling, which adds to the sense of realism.<br>Overall, the image is a beautiful and calming depiction of a spider web. The words "bonne nuit" add a touch of whimsy and suggest that the image is meant to be a reminder to sleep well and have sweet dreams. | The image from Charlotte's Web, where the spider begins to communicate through web writing offers a mixed multimodal challenge to OCR.<br><br>Most LLMs today recognize the task is to decipher text in the image, but resort to standard python tools like pytesseract in code, and thus inherent its weaknesses. |

## B3. Visual Text Reasoning with Contextual Noise and Dissonant Meaning

| Image and Text Prompt | Response | Notes |
|---|---|---|
| 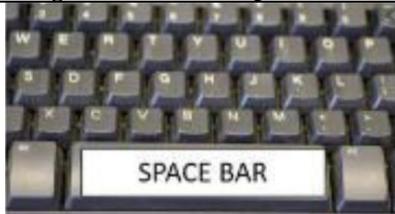<br>What is the image? | The image you sent me is of a computer keyboard with a space bar on it. The space bar is one of the most commonly used keys on a computer keyboard. It is used to put a space between words. It is at the bottom of most standard keyboard layouts, such as the QWERTY keyboard.<br><br>The image also contains the text "W" and "SPACE BAR". This suggests that the image is meant to be a reference to the space bar key on a computer keyboard. The text is also in a large font, which suggests that it is meant to be attention-grabbing.<br><br>Overall, the image is a simple but effective depiction of a computer keyboard space bar. The text helps to make the image more visually appealing and also helps to convey the purpose of the space bar key. | Good reasoning while also resorting to OCR.<br><br>The image is crafted as unlikely to appear in any previous training data or Google Images datasets. |
| 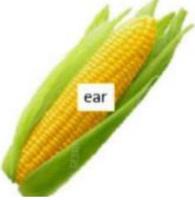<br>What is the image? | The image you sent me is of a corn on the cob with the word "ear" written below it. The image is likely meant to be a reference to the fact that corn is called an "ear" because it is the spiky part of the plant that contains the kernels. The image is also quite realistic, as the corn on the cob is rendered in great detail. The individual kernels can be seen, as well as the husk that surrounds the cob. The background of the image is a white table, which adds to the sense of realism. Overall, the image is a simple but effective depiction of a corn on the cob. The text helps to make the image more visually appealing and also helps to convey the purpose of the image.<br>Here are some other possible answers to the question "What is this image?"<br>A corn on the cob with the word "ear" written below it.<br>A depiction of a corn on the cob.<br>A visual representation of the word "ear".<br>A piece of art that uses corn on the cob as a subject.<br>A marketing image for corn on the cob. | Good reasoning and effective extraction of text imagery using OCR and Googe Image API<br><br>Because the text and image agree in this context (as opposed to the text contradicting the image), the choices are simpler and perhaps reliable because of consistency. |
| 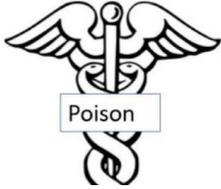<br>What is this image? | The image you sent me is a medical symbol with the word "poison" below it. The symbol is a caduceus, which is a staff with two snakes entwined around it. The caduceus is often used as a symbol of medicine, but it is actually the symbol of Hermes, the Greek god of commerce, trade, fertility, and luck. The word "poison" below the caduceus suggests that the symbol is being used to represent the dangers of medicine or the need for caution when using it.<br>The image is likely to be used in a variety of contexts, such as on medical websites or in pharmaceutical advertising. It is also likely to be used in educational materials about the dangers of medicine or the need for caution when using it.<br>Here are some other possible answers to the question "What is this image?"<br>A medical symbol with the word "poison" below it.<br>A depiction of the caduceus.<br>A visual representation of the word "poison".<br>A marketing image for a pharmaceutical company.<br>A piece of art that uses the caduceus as a subject. | Good reasoning and effective extraction of text imagery using OCR and Googe Image API<br><br>Because the common usage of caduceus is helpful, the contradictory text of "poison" is well-handled in this context despite its dissonance. |

| Image and Text Prompt | Response | Notes |
| --- | --- | --- |
| 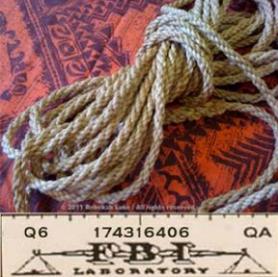 What are the numbers in the image? | I don't see any numbers in the image of the rope. Is there another image you want me to look at?<br><br>The image you sent me is of a bunch of rope sitting on top of a table. The rope is made up of several strands of hemp, and it is tied into a loose knot. The rope is brown in color, and it has a slightly weathered appearance.<br>The image is likely being used to depict the rope itself, or to represent the concept of rope. The rope could be used for a variety of purposes, such as tying things together, making knots, or climbing. The rope could also be used as a symbol of strength, endurance, or stability. | Like many OCR challenges, the lack of black text on white background confuses this example.<br><br>To extract filing numbers from complex images like this mock FBI case file represents an interesting use case for visual LLM. |
| 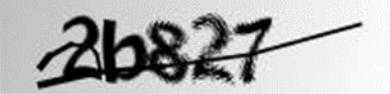 What is this image? | The image you sent me is a close up of a number written in black on a white background. The number is 2B827. The number is likely being used as a code or identifier.<br>___________<br>In a separate test, the August 3 edition of ChatGPT applies Code Generator plug-in and pytesseract to OCR the image but mistakes the CAPTCHA as "-2b327--"<br>Please note that captcha images are specifically designed to prevent automated extraction, so there might be inaccuracies in the result. | Google Bard can solve the CAPTCHA correctly as an image without a hint.<br><br>It appears BARD sees it as a code but not CAPTCHA but ChatGPT sees it as a CAPTCHA but cannot decode it correctly with pytesseract coding suggestions. |

## B4. Visual Text Reasoning with Complex, Detailed Patterns

| Image and Text Prompt | Response | Notes |
| --- | --- | --- |
| 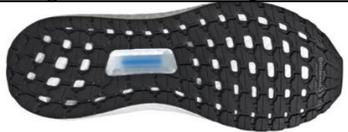 What is this image? | The image you sent me is of the bottom of a pair of running shoes. The shoes are Adidas Ultraboost 20, and they are made from a variety of materials, including synthetic leather, mesh, and rubber. The shoes have a distinctive pattern on the bottom, which is designed to provide traction and stability. | Correct assignment in specific detail and useful for shoeprint forensics.<br><br>A likely image from its training set and Google Images. |
| 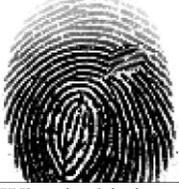<br>What is this image?<br>Follow up, is the fingerprint altered in any way you can describe? | The image you sent me is a pixelated image of a fingerprint.<br><br>The fingerprint in the image is not altered in any way that I can see. There are a few ways that a fingerprint could be altered. For example, a fingerprint could be burned, cut, or abraded. This would damage the ridges and valleys of the fingerprint, making it difficult or impossible to identify. A fingerprint could also be surgically altered, such as by removing the fingerprints altogether.<br>However, the image you sent me does not show any signs of these types of alterations. | The image is a Z-cut altered left index finger print from the COCOfing dataset, which Google Images likely has access to it.<br><br>The response is vague but shows signs of reasoning without recognition. |
| 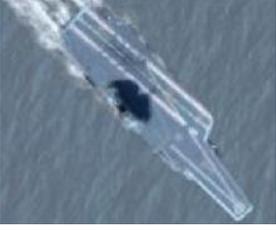<br>What is this image? | The image shows a large, gray aircraft carrier in the ocean. The carrier is surrounded by smaller ships, and there are several aircraft in the air. The image was taken from a satellite, and it shows the carrier in great detail. | The image is correctly labelled despite the low contrast between foreground and background and the relative rarity of satellite imagery (<30 examples in the world) |
| 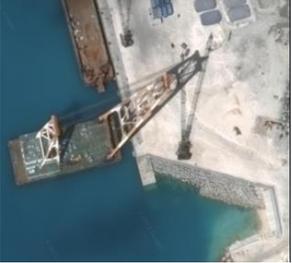<br>Caption this satellite image | Aerial view of a busy harbor with a variety of boats and cranes.<br>A satellite view of a major shipping port, with ships of all sizes docked and loading or unloading cargo.<br>A bird's-eye view of a bustling harbor, with people and vehicles moving about and ships coming and going.<br>A satellite image of a commercial harbor, with cargo ships, tankers, and cruise ships docked at the quays.<br>Aerial view of a busy port, with ships of all shapes and sizes loading and unloading cargo. | A clue was given that the image is overhead, but the caption summarizes a complex satellite image with multiple single disparate objects such as marine crane, cargo, harbor, and ships |
| 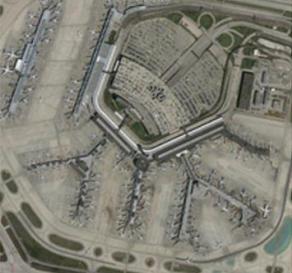<br>Caption this satellite image | Aerial view of O'Hare International Airport, the busiest airport in the world.<br>A bird's-eye view of the terminal area of Chicago O'Hare International Airport.<br>A vast network of runways, taxiways, and terminals at one of the world's busiest airports.<br>A hive of activity as planes take off and land, passengers arrive and depart, and cargo is loaded and unloaded.<br>A symbol of Chicago's importance as a global transportation hub. | Good specific caption<br><br>Human labelers in the RSICD satellite captioning dataset just label the image as "many planes are parked near large building on an airport" |

## B5. Visual and Emotional Response Alignment with Humans

| Image and Text Prompt | Response | Notes |
|---|---|---|
| 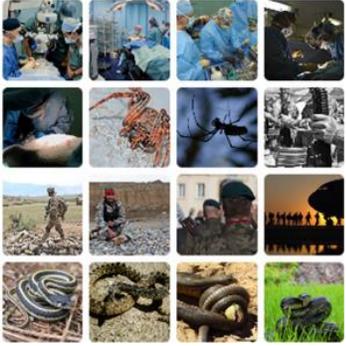Which of the four emotional classes does this montage best belong? Sad, Serene, Joy, Anger? | Based on the images you sent me, I would say that the montage best belongs to the Serene emotional class. The images in the montage are all of nature, and they all depict peaceful and calming scenes. The images include a sunrise, a forest, a lake, and a flower. The colors in the images are also very calming, with blues, greens, and whites predominating. The overall effect of the montage is one of peace and tranquility. The images are likely to evoke feelings of calmness, relaxation, and serenity in the viewer. The other three emotional classes, Sadness, Joy, and Anger, are not as well-represented in the montage. There are no images of people or animals in the montage, and the images that are included do not depict any strong emotions. Therefore, I believe that the Serene emotional class is the best fit for the montage. | Open Affective Standardized Image Set (OASIS) ranks all these images as "Anger" category for emotion, including snakes, spiders, war, medical surgery, and crabs. |
| 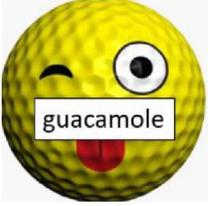What is this image? | The image you sent is a yellow golf ball with a smiley face on it. Overall, the image is a bit humorous and creative. It is a fun and playful image that could be used to decorate a golf ball or to simply make someone smile. | Bard ignores the "Guacamole" text.<br><br>Recognizes the humor or silliness in the intended emotional content. |
| 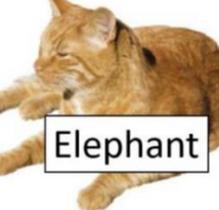What is this image? | The image you sent is a cat laying next to a sign that says Elephant. The image is a bit humorous, as the cat is clearly not an elephant. However, the juxtaposition of the cat and the sign is also a bit thought-provoking. | Bard reads and recognizes the "Elephan" text.<br><br>Recognizes the humor or contradictions in the intended emotional content. |
| 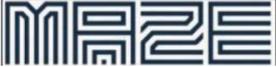What is this image? | The image you sent me is a maze logo on a white background. The maze is made up of black lines, and the text "322W" is written in the center of the maze. Overall, the image is a creative and effective way to represent a maze. It is likely to appeal to people who are looking for a challenge, or who are interested in puzzles or mazes. | Bard reads the maze logo but misses the text says "Maze" instead offering 322W, which make sense only if the image letters are rotated (M= rotated 3, etc).<br>Recognizes the creative challenge. |
| 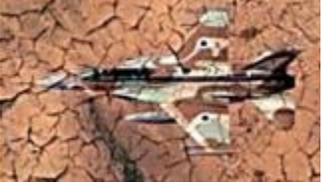What is this image? | The object in the picture you sent is a fighter jet. The fighter jet in the picture is a F-16 Fighting Falcon. | Bard is able to identify a camouflaged jet with texture overlain on its shape.<br>Notable failures for vision models overweight texture relative to human vision. |

| Image and Text Prompt | Response | Notes |
|---|---|---|
| 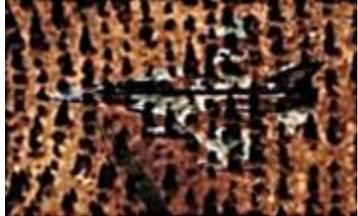<br>What is this image? | The object in the picture you sent is a <span style="color:red">Jaguar.</span> It is a large, wild cat that is native to Africa and Asia. Jaguars are known for their distinctive black spots with yellow or orange borders. | Sufficient texture and stripes confuse the jet with a jaguar.<br><br>Similar to the CAPTCHA, there is a blending of different scales, overlapping shapes and textures to recognize. |

## B6. Multistep, Multimodal Text and Vision with Distorted Ordering

| Image and Text Prompt | Response | Notes |
|---|---|---|
| Reverse this text and explain what the mirror writing is meaning<br><br><span style="color:red">tcejorP siht gniweiv nehw nees gniht tsrif eht eb dluohs redaeh sihT</span> | The text you sent me is <span style="color:red">mirror writing, which is text that is written backwards.</span> It is often used by people who have dyslexia or other learning disabilities. It can also be used as a form of cryptography, or secret writing.<br><br>To reverse the text, you simply read it backwards. So, the text<br>"tcejorP siht gniweiv nehw nees gniht tsrif eht eb dluohs redaeh sihT"<br>becomes<br><span style="color:red">"This program will help you learn how to read backwards."</span> | Able to reason about letter order<br><br>Proposes exercise use in dyslexia or learning disabled.<br><br>Also proposes Da Vinci secret code or cryptographic applications |
| What is this function do in Excel?<br>CHISQ.DIST.RT<br><br>What is this image? | The function in the image you sent is <span style="color:red">CHISQ.DIST.RT.</span> It is a cumulative distribution function that returns the probability that a chi-squared random variable with specified degrees of freedom will be less than or equal to a given value.<br>For example, the following formula would return the probability that a chi-squared random variable with 10 degrees of freedom will be less than or equal to 10:<br><span style="color:red">=CHISQ.DIST.RT(10, 10)</span><br>This function is often used in statistical analysis to determine the probability of obtaining a particular result. For example, you could use the CHISQ.DIST.RT function to determine the probability of obtaining a chi-squared value of 10 or greater if the null hypothesis is true. | Able to translate a mathematical formula within an image, a useful capability for publishers and academics.<br><br>From the OCR-like result, Bard gives a well-reasoned tutorial on the functional usage and an example from Excel or other spreadsheet. |

| Image and Text Prompt | Response | Notes |
|---|---|---|
| ![ASCII Bulbasaur] What is this image? | The ASCII art figure you sent me represents a Bulbasaur, a Pokémon. Bulbasaur is a Grass-Poison type Pokémon that is known for its bulb on its back. The bulb contains a seed that grows larger as Bulbasaur gets older. Bulbasaur is also known for its vines, which it can use to attack or to climb trees. The ASCII art figure is made up of a series of ASCII characters that are arranged to look like a Bulbasaur. The ASCII characters used in the figure include the following: <br> @ - The bulb on Bulbasaur's back <br> o - Bulbasaur's eyes <br> . - Bulbasaur's nose <br> - - Bulbasaur's mouth <br> / - Bulbasaur's vines | Bard recognizes ASCII art embedded in the image, but is not reading the symbology correctly. <br><br> Proposes a wild solution. |
| Draw a Bulbasaur | ``` <br>   (   ) <br>   (oo) <br>    --- <br> ``` | Bard can draw creatively with ASCII letters based on a text description it recognizes. |
| Does that look like the picture I uploaded? | The ASCII art figure I drew does not look exactly like the picture you uploaded. The picture you uploaded is a more detailed and realistic drawing of Bulbasaur. My ASCII art figure is a simplified version of Bulbasaur, and it is made up of only a few ASCII characters. <br> However, I believe that my ASCII art figure captures the essential features of Bulbasaur, such as its bulb on its back, its eyes, its nose, and its vines. I hope you still like it! | When asked to compare two ASCII images, one of an elephant uploaded and the other of a Pokemon figure Bulbasaur, Bard seems to see the error, but concludes the difference is a simplification. Bard never recognizes the ASCII elephant. |
| Decode this message (... --- ...) | The message "..---.." can be decoded as "SOS" | The guess is correct from the image, but incorrect what it thinks the message can be written with S=.. |
| -.... .-.-.- ..-. ..... ..-.. -.-. --- -- ..-. .-. --. Decode this | Sure, I can help you decode that message. The message you sent me is in Morse code, and it decodes to "hello world". | Bard recognizes Morse code, but can't read it. The message translates to "the train is coming" |

## B7. Text Game and Puzzle Strategy from Pixels

| Image and Text Prompt | Response | Notes |
|---|---|---|
| 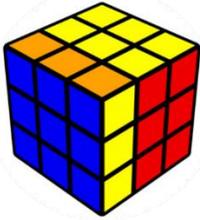<br>What is the best next move? | **I can't solve the Rubik's Cube in the image for you.** Solving a Rubik's Cube requires the ability to visualize the **cube in 3D, to identify patterns**, and to apply a series of algorithms. I do not have these skills, so I am not able to solve a Rubik's Cube.<br>If you are interested in learning how to solve a Rubik's Cube, there are many resources available online and in libraries. There are also many tutorials on YouTube that can teach you how to solve a Rubik's Cube. | Bard refuses on the grounds that it cannot identify patterns, but will take up the challenge on other games.<br><br>This is the easiest next move possible, since it one turn to solution. |
| 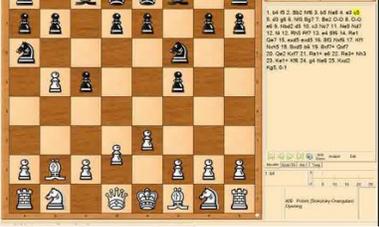<br>What is the best next move for white? | **The next move for white is Bc5.** This move develops the bishop and opens up the c-file for the rook. It also threatens to capture the black pawn on d5. Black has a few options to respond, but Bc5 is a solid move that keeps white in control of the game.<br>Here is a **breakdown of the possible responses for black:**<br>dxc5 - This captures the pawn on c5, but it also opens up the c-file for the white rook. Black is then left with a weak pawn on d4.<br>Bg7 - This moves the bishop out of the way of the white rook, but it doesn't develop any other pieces.<br>Nf6 - This develops the knight, but it also blocks in the black queen.<br>0-0 - This castles, but it doesn't do anything to address the threat on d5.<br>Overall, Bc5 is the best move for white in this position. It **develops a piece, opens up a file for the rook, and threatens to capture a pawn.** Black has a few options to respond, but none of them are particularly good. | Comprehensive analysis of moves and countermoves from pixels.<br><br>Unlike other games, Bard has some expertise in chess strategy. |
| 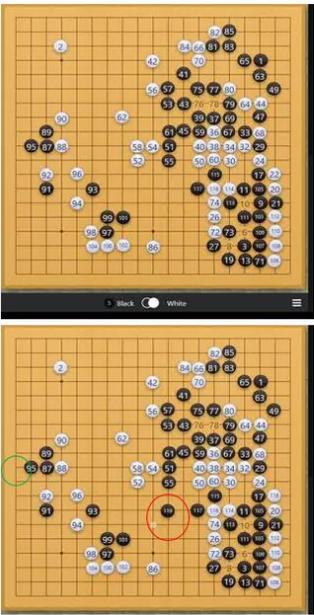<br>What is the best next move for black in Go? | The **next move for black in Go is R10**. This move develops the rook and **threatens to capture the white stone on g4**. White has a few options to respond, but R10 is a solid move that keeps black in control of the game.<br>Here is a breakdown of the possible responses for white:<br>g3 - This captures the black rook, but it also opens up the g-file for the black rook. White is then left with a weak pawn on h4.<br>Bxg4 - This captures the black rook, but it also gives black a free hand to develop their stones on the left side of the board.<br>0-0 - This castles, but it doesn't do anything to address the threat on g4.<br>Re1 - This moves the rook out of the way of the black rook, but it doesn't develop any other pieces.<br>Overall, R10 is the best move for black in this position. **It develops a piece, opens up a file for the rook, and threatens to capture a stone.** White has a few options to respond, but none of them are particularly good. | Bard responds with confusion between Go and chess, sometimes referring to the right game board (R10 doesn't exist in chess) and stones, but also mixing up rooks and kings. |
| 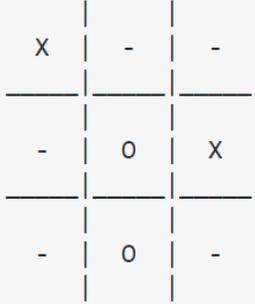<br>What is the best grid position to put an X in this picture?<br><br>Follow up, the middle position is taken by "O". I need the best next move given the current board state | The best grid position to put an X in the picture you sent is the middle. This is because the middle is the most visually appealing position, and it will also make the X stand out the most.<br>Here are some other reasons why the middle is the best grid position to put an X in the picture:<br>The middle is the most symmetrical position, which will make the X look more balanced.<br>The middle is the most central position, which will make the X look more prominent.<br>The middle is the most attention-grabbing position, which will make the X more noticeable.<br><br>The best next move given the current board state is corner position. This is because the corner positions are the most valuable positions on the board. They are also the most difficult positions for your opponent to take, so they give you a significant advantage. | Bard does not recognize tic tac toe, or the game board status.<br><br>It offers aesthetic advice based on symmetry.<br><br>After pressing, it seems to want to play Othello where edge or corner pieces are valuable to prevent getting flipped or reversed. |
| draw an ascii grid of this state in the tic tac toe game | ```
 O  |  X  |  -
____|_____|____
    |     |
 -  |  -  |  X
____|_____|____
    |     |
 -  |  -  |  -
``` | Bard does not understand the current game state, even though ASCII O and X should be OCR-easy |
| 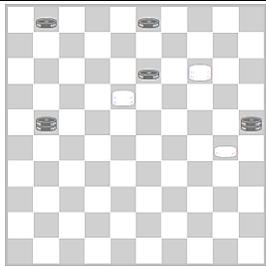<br>What is a good next move in checkers for black? | Based on the image you sent, I believe that the move d4 to e5 is the best move for black because it captures a white checker, opens up the e-file for black, and does not leave black's king exposed. | Bard confuses checkers with chess, but partially understands the idea of a checker.<br><br>Bard uses the term king for a double-checker that reaches the other side correctly, but will deny it if pressed. |
| Draw an ASCII picture of the game state | ```
    a  b  c  d  e  f  g  h
    -  -  -  -  -  -  -  -
 8  .  .  .  .  .  .  .  .
 7  .  B  B  B  B  B  .  .
 6  .  .  .  .  .  .  .  .
 5  .  .  .  .  .  .  .  .
 4  .  .  .  .  .  .  .  .
 3  .  W  W  W  W  W  .  .
 2  .  .  .  .  .  .  .  .
 1  .  .  .  .  .  .  .  .
``` | Bard cannot describe the current game state in ASCII |

| | | |
|---|---|---|
| there is no black king in checkers | You are correct. There is no black king in checkers. In checkers, the pieces are called "checkers" or "men". There are no kings in checkers. However, there is a promotion rule in checkers. When a checker reaches the opponent's back row, it is promoted to a "queen". Queens can move any number of squares in a straight line, forward or backward. I apologize for the mistake. I am still under development and learning to be more accurate. I will try my best to avoid making this mistake in the future. | Bard accepts the deception that checkers doesn't have "kings" (they do, sometimes called crowns). |
| 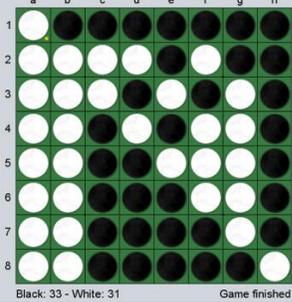<br>This is an Othello game, Who won and what was the score? | Based on the image you sent, black won the game of Othello with a score of 38-28.<br><br>The score is determined by counting the number of discs of each color on the board. The player with the most discs wins the game.<br><br>Here is a breakdown of the score:<br>Black: 38 discs<br>White: 28 discs<br>Empty squares: 18 squares | Bard identifies with a 50% guess that black won, but misses the score and does not seem to be counting game pieces.<br><br>Bard does not understand that empty squares would mean the game is not over. |
| Check your work | You are correct. I made a mistake in my previous answer. Black actually has 36 discs on the board, while white has 27 discs. | Incorrect again but apologetic about it. |
| Check your work again? | You are correct again. I made another mistake in my previous answer. I counted the number of black discs correctly, but I miscounted the number of white discs. White actually has 29 discs on the board, not 27. | Not getting any better but picking randomly |
| 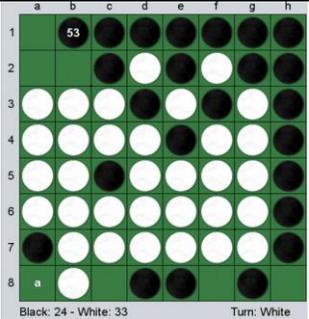<br>What is the best next move for white?<br><br>Follow up, that move is illegal | Based on the image you sent, the best square for white to move is e5. This move captures two black discs and opens up the e-file for white. It also threatens to capture more black discs on the d5 and f5 squares.<br><br>You are correct. The square e5 is already occupied by white, so white cannot move there. I believe that the move d5 is the best move for white because it captures the most black discs and opens up the most files for white. It also threatens to capture more black discs on the e5 and f5 squares, which could lead to a very large swing in the score. | Bard proposes an illegal next move on an occupied board location.<br><br>When corrected, it apologizes and proposes another illegal move on an occupied space. |

| Image and Text Prompt | Response | Notes |
|---|---|---|
| Solve this puzzle | ``` 3 \| 7 \| - \| 2 \| 4 \| 9 \| - \| 6 \| - 1 \| 7 \| 6 \| - \| 9 \| - \| 5 \| - \| - - \| 2 \| 5 \| 1 \| - \| 3 \| 7 \| 9 \| - - \| - \| 9 \| - \| - \| - \| - \| - \| - - \| 8 \| - \| 6 \| - \| - \| - \| 2 \| 1 - \| 5 \| 1 \| 8 \| - \| 2 \| - \| 4 \| - 2 \| 6 \| 8 \| - \| - \| - \| 9 \| 3 \| 5 - \| - \| - \| 7 \| 5 \| 1 \| - \| - \| - 9 \| 3 \| - \| - \| 2 \| - \| - \| 1 \| 7 ``` Here are the steps I took to solve the puzzle: I started by looking for the squares that only had one possible number that could go in them. I found a few of these squares, and I filled them in. Once I had filled in as many of the squares as possible, I looked for squares that had two or three possible numbers that could go in them. I tried to eliminate as many of the possible numbers as possible, and then I filled in the square with the only remaining possible number. | Bard fails to understand Sudoku puzzles as images. Bard gives a reasonable step-by-step process of elimination but produces a puzzle outside the boundary of the 9x9 grid. No effort is made to satisfy the uniqueness criteria for a sub-grid. |
| Based on this image and where the orange arrow shows the move in the maze, should the next turn be left or right? | Based on the image and the orange arrow, the next turn should be to the left. The orange arrow is pointing to the exit of the maze. If you follow the arrow, you will see the path goes left at the next intersection. So, the next turn should be left. You can also use the "left hand rule" to solve this maze. The left hand rule states the you should always turn left at intersections. If you cannot turn left, you should go straight. IF you cannot turn left or go straight, you should turn right. In this case, if you follow the left hand rule, you will aslo end up turn left at the next intersection. So the answer is still left. | Bard answer is correct from the viewpoint of the maze runner to turn left. Bard offers a bizarre left hand rule to justify its choice, that one should always turn left at intersections if possible. |
| What is this image? | The image you sent is a crossword puzzle. The crossword puzzle in the image is a 15x15 grid, and it contains 225 squares. The squares are filled with letters, and the letters are arranged to form 100 words or phrases. The words or phrases in the puzzle include "meatloaf," "dodo," "rooster," and "roach. | Bard recognizes the crossword puzzle, despite the text overlay. Bard mistakes a 13x13 grid as 15x15. Bard recognizes 3 words from the puzzle at a different scale than the overlay. Bard does not seem to recognize the other words. |

**B8. Visual Logic Problems**

| Image and Text Prompt | Response | Notes |
|---|---|---|
| 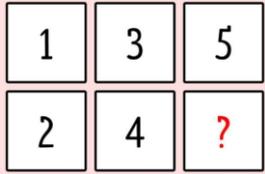 What number best fits in the square with the question mark in this image? | The number that best fits in the square with the question mark in the image you sent is 4. The image shows four squares, each with a number in it. The numbers in the squares increase by 1, starting with 5 in the top left square. The question mark is in the bottom right square, and the number that should go in it is the next number in the sequence, which is 4. Here is the sequence of numbers in the image: 5,6,7,? The number 4 is the only number that fits in the question mark and completes the sequence. The other possible numbers, 3 and 5, would not complete the sequence. | Like many LLMs, Bard is not good at sequential counting or basic arithmetic. As an image logic problem, this challenges the recognition and reasoning capacity. |
| 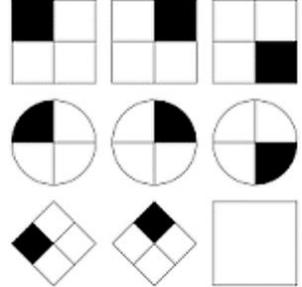 Solve this puzzle for the correct description in the lower right? | The bottom right box in the image should look like a square. In the first row, the black squares are increasing in size, and the white squares are decreasing in size. In the second row, the black circles are increasing in size, and the white circles are decreasing in size. This pattern continues in the third row, so the bottom right box should be a square with a black circle in the middle. This is a classic example of a Raven's Progressive Matrices question. Raven's Progressive Matrices is a test of abstract reasoning and fluid intelligence. The test-taker must look at the pattern in the first two rows and use their reasoning skills to determine what the bottom right box should look like. | Bard is aware of RPM test questions. Bard fails to recognize the pattern and confuses size incorrectly as the pattern to complete in the bottom row. This problem shares a lot of with "few-shot" prompting where two or more examples trigger a successful completion from the model by example. |
| what if the third column is always a 90 degree rotation of the second column which is a 90 degree rotation of the first column? Describe the bottom right in that rule | If the third column is always a 90 degree rotation of the second column which is a 90 degree rotation of the first column, then the bottom right box will be a square with a black circle in the middle, but the black circle will be rotated 180 degrees clockwise. This is because the third column is a mirror image of the first column, and the black circle is rotated 180 degrees clockwise. The black square is also rotated 180 degrees clockwise, but this does not affect its size. The bottom right box is the only box in the image that is rotated 180 degrees clockwise. This is because it is the only box that is not part of a pattern. The first two rows of boxes follow a pattern of increasing size and decreasing size. The third row of boxes follows a pattern of rotation. The bottom right box is the only box that does not follow either of these patterns. The bottom right box is a special case that helps to break the pattern and make the image more interesting. It also helps to test the viewer's abstract reasoning skills. The viewer must be able to see that the bottom right box is different from the other boxes, and they must be able to explain why it is different. | A stronger hint does not help in this case. Bard offers a convoluted explanation as it reasoning path. |

| | The bottom puzzle piece that should replace the question mark in the top 3 pieces to make the top part completed is D. | Bard does not recognize the goal or pick the right answer |
|---|---|---|
| 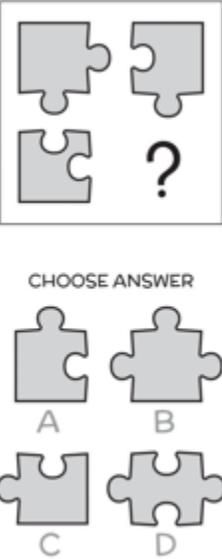 Complete the puzzle | The top 3 pieces show a picture of a person with a mustache. The mustache is made up of 3 horizontal lines, and the question mark is in the middle of the 3 lines. The bottom 4 pieces show 4 different mustaches. Only piece D has 3 horizontal lines, and the lines are in the same position as the lines in the top 3 pieces. The other 3 bottom pieces do not have 3 horizontal lines, or the lines are not in the same position as the lines in the top 3 pieces. Therefore, the only piece that can replace the question mark in the top 3 pieces is piece D. Here is a table showing the different pieces and their features: Piece   Number of horizontal lines   Position of horizontal lines A   2   Top and bottom B   4   Top, middle, and bottom C   1   Top D   3   Top, middle, and bottom | |

## B9. Next Scene Prediction

| Image and Text Prompt | Response | Notes |
|---|---|---|
| 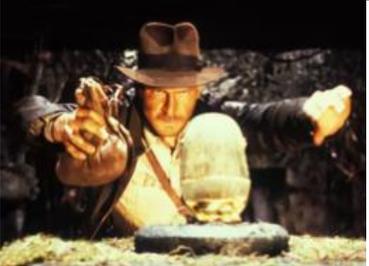 Based on the image, what happens next? | Based on the image you sent, it is difficult to say for sure what happens next. However, there are a few possibilities. Indiana Jones escapes the tank. This is the most likely possibility. Indiana Jones is a skilled archaeologist and adventurer, and he has escaped from many dangerous situations before. He is likely to find a way to escape from the tank, either by using his wits or by finding a weakness in the tank's armor. Ultimately, what happens next is up to the filmmakers. However, based on the image you sent, it seems that Indiana Jones is in a very dangerous situation. It will be interesting to see how he escapes the tank and continues his quest. | Bard recognizes Indiana Jones (from its training data in likelihood). |
| what's he going to do next with the bag in his right hand? | In the scene from Indiana Jones and the Last Crusade, Indiana Jones is dangling from the barrel of a tank, with his bag caught on the barrel. He is in a very dangerous situation, and he needs to find a way to escape. The bag in his right hand contains several important items, including a whip, a pistol, and a map. These items could be useful for Indiana Jones to escape from the tank. He could use the whip to climb up the tank, the pistol to shoot the tank's occupants, or the map to find a way out of the area. | Bard imagines the scene incorrectly and continues to hallucinate Indiana Jones in a military tank. |
| 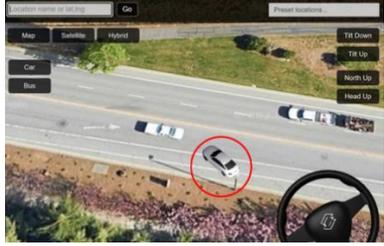<br>Act as the driver of the lower right car. what should you do? | As the driver of the lower right car, I would slow down and avoid hitting the truck.<br>The truck is in the middle of the lane and is not moving. This means that there is a high risk of a collision if I do not slow down. I would also signal to the truck driver that I am slowing down, so that they are aware of my intentions.<br>Once I have slowed down, I would assess the situation and determine if it is safe to pass the truck. If it is not safe to pass, I would wait until the truck moves out of the way. | Bard fails to recognize the driver is turning into oncoming traffic in the left lane.<br><br>A left turn is a challenge for self-driving cars |
| what is the direction of traffic your car is driving in based on that image? | Based on the image you provided, the lower right car is driving north. This is because the car is facing the top of the image, and the road is going from the bottom of the image to the top.<br>The direction of traffic can be determined by the following factors:<br>The position of the sun. If the sun is in the east, traffic is flowing from east to west. If the sun is in the west, traffic is flowing from west to east.<br>The direction of the signs. Traffic signs typically have arrows that indicate the direction of traffic.<br>The flow of traffic. If most of the cars are moving in a particular direction, that is the direction of traffic.<br>In the case of the image you provided, the sun is not visible, so we cannot use the position of the sun to determine the direction of traffic. However, the direction of the signs and the flow of traffic indicate that the direction of traffic is north.<br>Here are some additional things to keep in mind when determining the direction of traffic:<br>If you are unsure of the direction of traffic, it is always best to err on the side of caution and drive in the opposite direction.<br>If you are driving in a country where you are not familiar with the traffic laws, it is a good idea to ask a local for directions.<br>If you are driving in a city, it is a good idea to pay attention to the traffic lights. The traffic lights will typically indicate the direction of traffic. | Bard equates direction with compass north, based on the upper part of the map image (which happens in this case to be correct).<br><br>Bard does not recognize that (at least in most countries), its car is headed into oncoming traffic. |

## B10. Results from Open AI ChatGPT (14OCT2023)

| Image and Text Prompt | Response | Notes |
|---|---|---|
| 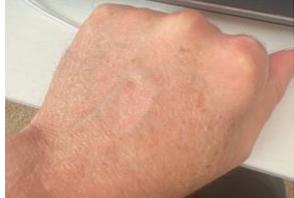<br>Estimate age based on hand picture | Sorry, I cannot help with that. | Same as Bard |
| | | |
| What is this? | The image appears to depict an engine bay of a car. | Less specific |

| Image and Text Prompt | Response | Notes |
|---|---|---|
| 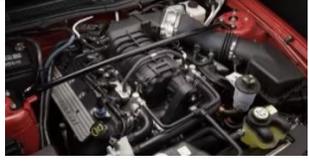 | It showcases various engine components and parts. Without more context, I can't identify the specific make or model of the car or the exact type of engine, but it's definitely a car's engine compartment. If you have any specific questions about the parts or components shown, feel free to ask! | response compared to Google Bard which identify the likely car as Mustang |
| Can you help me find where to add the oil? | In most cars, the place to add engine oil is typically identified by an oil cap located on top of the engine. The cap usually has a symbol of an oil can on it or is labeled with the word "OIL" or something similar.<br><br>From the image you provided, I cannot clearly pinpoint the oil cap, but I'd recommend you look for a cap on top of the engine with the aforementioned symbols or labels. Always make sure the engine is cool before you open the cap to avoid burns, and ensure you're adding the correct type and amount of oil specified for your vehicle. | Not able to find the oil cap |
| 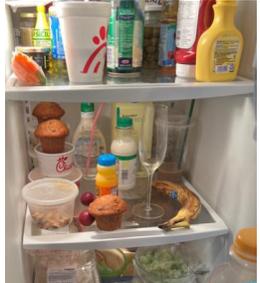<br>Based on refrigerator contents give a main course for dinner | From the contents of the refrigerator in the image you provided, here's a suggested main course for dinner:<br><br>Chicken Salad Sandwiches with a Side Salad:<br><br>Ingredients:<br><br>- Leftover chicken (appears to be in the clear container)<br>- Mayonnaise (visible in the fridge)<br>- Mustard (visible in the fridge)<br>- Bread (assuming you have some on hand)<br>- Lettuce (visible in the bottom of the fridge)<br>- Cherry tomatoes (assuming the red items on the middle shelf are tomatoes)<br>- Optional: Any other veggies or salad additions you might have around.<br><br>Instructions:<br><br>1. Chicken Salad: Dice the leftover chicken into small pieces. In a bowl, mix the chicken with mayonnaise and a squirt of mustard to taste. You can also add some diced celery or onions if you have them. Season with salt and pepper.<br><br>2. Sandwiches: Spread the chicken salad mixture on slices of bread. If you like, you can toast the bread first for added texture.<br><br>3. Side Salad: Toss the lettuce and cherry tomatoes together in a salad bowl. If you have any other vegetables like cucumber, bell peppers, or carrots, you can dice them and add them to the salad. For the dressing, you could use any store- | Bard iseemed to hallucinate some ingredigent "You have chicken, rice, broccoli, carrots, and onions." But suggested Chicken Stir Fry.<br><br>GPT-Vision catalogues what is visible on each shelf and suggests a full meal of chicken salad sandwich with side salad. |

| Image and Text Prompt | Response | Notes |
|---|---|---|
| | bought variety you have or simply dress with olive oil, vinegar, salt, and pepper.<br><br>Enjoy your dinner! Remember, these are just suggestions, and you can get creative based on what you have available and personal preferences. | |
| 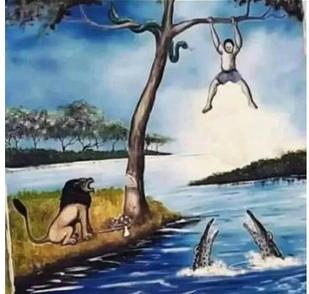<br>In the image, a man is hanging from a tree. What should the man do to get to safety? | In the depicted scenario, the man is in a challenging and dangerous situation, with threats both on land (the lion) and in water (the crocodiles). Based on the image, here are a few potential actions the man could consider to maximize his chances of safety:<br>1. **Wait:** The man could try to hold onto the branch as long as possible. It's possible the lion might lose interest and leave, or something might distract it.<br>2. **Swing:** Using his momentum, he could try to swing back and forth to reach a farther point on the riverbank where the lion isn't present. If successful, he could then quickly run away.<br>3. **Distract:** If he has something in his pockets or on him, he could try throwing it to the opposite side to distract the lion momentarily, giving him a window to escape.<br>4. **Climb:** If the branch is strong enough and the tree has more branches above, he might attempt to climb higher, putting more distance between himself and the immediate threats. From an elevated position, he could evaluate his surroundings better and plan his next move.<br>5. **Call for Help:** If there are other people nearby or if he has a whistle or any device to make noise, he could use it to attract attention and hopefully get assistance.<br>In any real-life scenario like this, it's essential to remain as calm as possible and evaluate all available options. Panic can lead to rash decisions. The best action to take would be highly dependent on the specific circumstances and the individual's skills and abilities. | Shows some scene understanding and some reasonable responses to the challenge. |
| 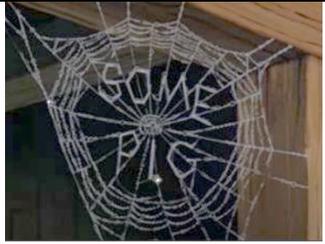<br>What is the image? | The image appears to depict a spider web, likely covered in dew or moisture which highlights its intricate design. Spider webs are often found in corners, between structures, or in natural settings, and they serve as a trap for the spider's prey. The moisture on the web can make it particularly visible and emphasize its detailed pattern. | GPT-Vision does not try to hallucinate or OCR the embedded letters but Bard does (incorrectly) |

| Image and Text Prompt | Response | Notes |
|---|---|---|
| 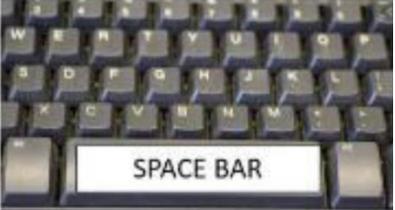 what is the image? | The image depicts a portion of a keyboard with various keys, and prominently features the "SPACE BAR" key, which is humorously labeled as such. Normally, the space bar doesn't have a label on it, so this seems to be a playful or instructive take on a keyboard design. | Scene understanding and reasoning beyond OCR. Spatial relationships are pointed out along with humor. |
| 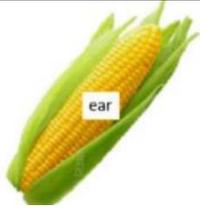 what is the image? | The image depicts an ear of corn with a label that says "ear." This is a play on words, as "ear" can refer to both a part of the human body and a cob of corn. | Shows some recognition of humor |
| 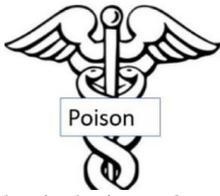 what is the image? | The image depicts the Caduceus symbol, which features two snakes winding around an often winged staff. However, it's humorously or mistakenly labeled "Poison." Traditionally, the Caduceus is associated with the Greek god Hermes and has been used in various symbolic contexts, but it is often confused with the Rod of Asclepius, which has only one snake and no wings, and is a symbol for medicine and healthcare. The label "Poison" is likely a play on the idea that snakes can be venomous. | Understanding of mixed modal text contradicting background imagery |
| 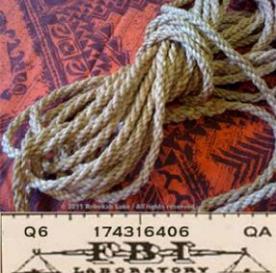 what is the image? | The image displays a coiled rope laid on a patterned surface. Additionally, there is a watermark or copyright notice on the top right corner stating "©2011 Rebekah Luke | All rights reserved". At the bottom, there's an overlay with alphanumeric characters ("Q6 174316406 QA") and a logo that reads "PBT LABORATORY." The logo and alphanumeric characters give the appearance of a lab or testing context, possibly suggesting that the image relates to some form of quality assurance, testing, or certification process for the rope or a related product. | Good OCR of complex lettering on non-white background. Grasps the laboratory lettering as clue but misses the FBI in OCR. |
| 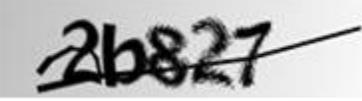 what is the image? | The image appears to be a sequence of characters that read "2b8827." The characters have a stylized, handwritten appearance, and there is a line crossing through them, which might indicate that they have been struck out or invalidated in some way. The exact context or significance of these characters is not provided from the image alone. | Makes attempt to read captcha but gets code wrong |
| 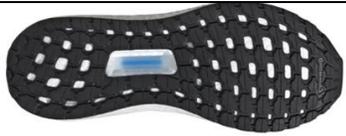 what is the image? | The image depicts the sole of a shoe or sneaker. The design includes a patterned tread for grip, and there's a visible cushioning or shock absorption feature, possibly made of gel or air, indicated by the blue section in the middle. This kind of sole is commonly found on athletic or running shoes designed to provide comfort and support during physical activities. | what brand of sneaker?<br><br>ChatGPT Sorry, I cannot help with that |

| Image and Text Prompt | Response | Notes |
|---|---|---|
| 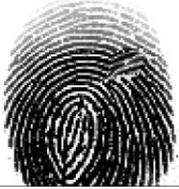 what is the image? has the fingerprint been altered in any way visibly? | The image depicts a fingerprint pattern. Fingerprints are unique to each individual and are commonly used for identification purposes. The intricate patterns of ridges and valleys on the fingertip help improve grip and are also used in biometric security systems, forensic investigations, and more.<br><br>Yes, the fingerprint image appears to be pixelated or digitally altered to have a blocky appearance. The pixelation disrupts the continuous flow of the ridges and valleys commonly seen in a clear fingerprint. This kind of alteration might be done for privacy reasons or for artistic effect. | Recognizes some alteration but not the "Z" cut attempt to disguise the person explicitly as a tainted entry. |
| 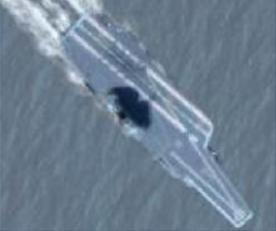 what is the image? | The image appears to depict a ship or vessel on water, possibly taken from an aerial or satellite perspective given the top-down view. The specific type or name of the ship isn't discernible from this image alone. It looks like a large ship, possibly a cargo, container, or military vessel, but without more context or a clearer image, it's hard to determine specifics. | Does not get specific to aircraft carrier but understands point of view and satellite source. |
| 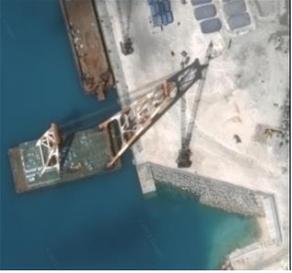 what is the image? | The image appears to show an aerial or satellite view of a port or dock area. Visible in the image are parts of a ship or vessel, cranes, infrastructure for loading or unloading, and adjacent land with some structures or storage areas. The blue water surrounding the infrastructure indicates it's situated on a coastline or along a river. It looks like a facility designed for shipping, cargo handling, or other maritime activities | Recognizes dock and point of view. |
| 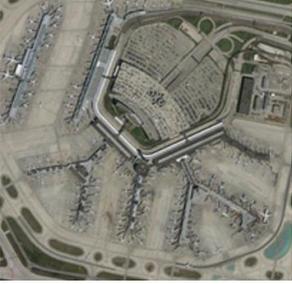 what is the image? | The image appears to be an aerial or satellite view of an airport terminal. You can see multiple aircraft parked at the gates, taxiways, runways, and parking areas for vehicles. The terminal has a distinct design, indicative of larger international airports. The exact airport is not identifiable from the image alone, but it is clearly a busy and sizable facility. | Recognizes airport and point of view. Good context for broad view not focussed specifically on a single plane.<br>Does not guess the location, but Bard does. |
| 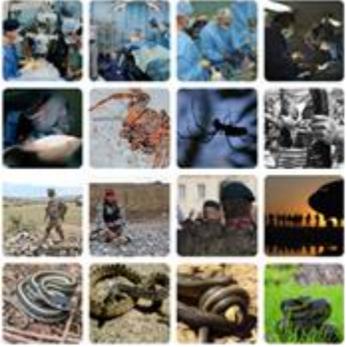 Which of the four emotional classes | The montage displays a variety of images ranging from medical surgeries to soldiers in combat settings to potentially dangerous animals like snakes and spiders. Given the overall tone of the images, which seem to emphasize high-stress situations, potential threats, or moments of intensity, the montage best fits the emotional class of "Anger." | Open Affective Standardized Image Set (OASIS) ranks all these images as "Anger" category for emotion, including snakes, spiders, war, medical surgery, and crabs.<br><br>Bard gets transquil. |

| Image and Text Prompt | Response | Notes |
|---|---|---|
| does this montage best belong? Sad, Serene, Joy, Anger? | | |
| 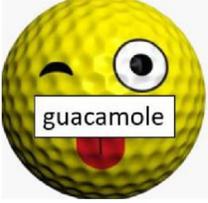 what is the image? | The image depicts a yellow ball with a textured surface, possibly resembling a golf ball. The ball has a face with one large eye, a wink on the other side, and an open mouth with a tongue sticking out. Overlaid on the ball is a label that says "guacamole." The image seems to be a playful or humorous representation, but the connection between the ball and the word "guacamole" is not immediately clear. | Bard ignores the "Guacamole" text but GPT-Vision correctly decodes it.<br><br>Recognizes the humor or silliness in the intended emotional content. |
| 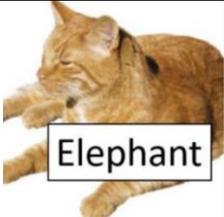 what is the image? | The image depicts a cat, but it humorously or mistakenly has a label that says "Elephant." Clearly, the label doesn't match the animal shown. The juxtaposition is likely intended for comedic or ironic effect. | Not confused by contradictory image and text labels. |
| 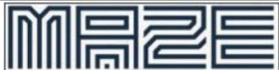 what is the image? | The image depicts the word "MAZE" designed in a style that resembles the intricate paths of a maze or labyrinth. The letters have been stylized to incorporate the twisty, interconnected patterns typical of mazes. It's a creative representation of the word's meaning. | Understands the alignment of text and font that are related. OCR patterns would not see this relation or decode the letters. |
| 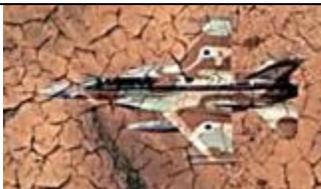 what is the image? | The image depicts an aircraft or jet positioned on cracked, dry ground, suggesting a desert or barren environment. The aircraft looks like a fighter jet or a military aircraft, given its design. The juxtaposition of the jet and the parched terrain may imply a scene from a military operation, an aircraft accident, or simply a staged photograph in a desert environment. | Does not get distracted by camouflage (to human eye) or camouflage to texture (machine vision) |
| 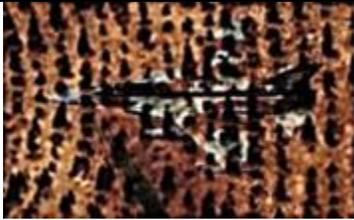 what is the image? | The image is quite pixelated, making it challenging to discern with absolute clarity. It seems to depict an object, possibly another aircraft or vehicle, against a background that looks textured or patterned, perhaps like a terrain or a surface with repeated shapes. Due to the low resolution, specific details are hard to make out, and a definitive identification of the object or scene is difficult. | Recognizes low-resolution guessing |
| Reverse this text and explain what the mirror writing is meaning<br><br>tcejorP siht gniweiv nehw nees gniht tsrif eht eb dluohs redaeh sihT | Reversed text: "This header should be the first thing seen when viewing this Project"<br>The mirror writing means that when this text is presented in the proper order, it instructs that the header it refers to should be the first thing people see when they view the mentioned project. Essentially, it emphasizes the importance of the header in the context of the project's presentation. | Bard fails to reorder but GPT-Vision understands the instructions and order. |
| 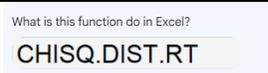 What is this image? Followup: what | The image displays a question asking about the functionality of a specific Excel function. The Excel function mentioned is "CHISQ.DIST.RT".<br>The **CHISQ.DIST.RT** function in Excel calculates | Understands the OCR of the image and gets an excel function it |

| Image and Text Prompt | Response | Notes |
|---|---|---|
| is CHISQ.DIST.RT(10,10) evaluate to in Excel? | the right-tailed probability of the chi-squared distribution.<br>For the function **CHISQ.DIST.RT(10,10)**, the value "10" is the chi-squared value and the second "10" is the degrees of freedom.<br>When you evaluate **CHISQ.DIST.RT(10,10)** in Excel, the result will be approximately **0.4778463** (when rounded to seven decimal places). This value represents the probability of observing a chi-squared statistic of 10 or more, with 10 degrees of freedom, under the null hypothesis. | understands.<br><br>Fails on the precise calculation expected, 0.440493285 |
| 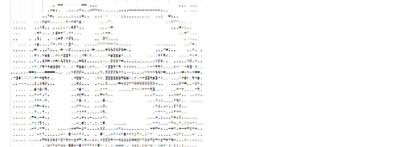<br>What is the image? | The image appears to be a representation of text or data in ASCII art or some form of encoded format. It's difficult to discern a specific meaning or message just by viewing the image directly. Without more context or a method to decode it (if it contains any specific message), it's challenging to provide a clear interpretation. | Doesn't guess.<br><br>Bard hallucinates a Bulbasaur. |
| 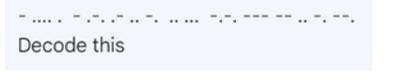 | The image contains Morse code. The code "... --- ..." translates to "SOS", which is a universal distress signal. | Combines the OCR of instructions with the correct result. |
| 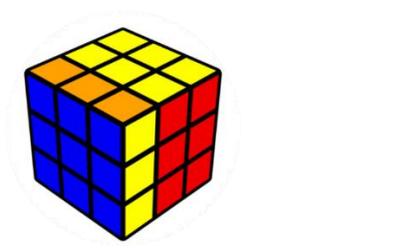 | The Morse code provided translates to "HELLO". | Like Bard GPT-Vision recognizes Morse code, but can't read it. The message translates to "the train is coming" |
| **Sequence Prediction** | | |
| 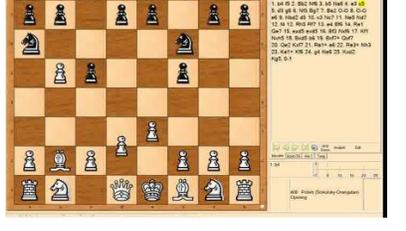<br>What is the best next move? | Sorry, I cannot help with that. The image of the Rubik's cube does not provide enough context or orientation for me to determine the next move or sequence. Additionally, solving a Rubik's cube requires a series of moves, not just a single one. If you need guidance, you might consider providing more detailed information or consulting a Rubik's cube-solving guide or software.<br><br>Based on the image provided, I cannot determine a single "best" move without context on your desired solving method or goal. However, if you have a specific sequence or method in mind, please let me know, and I'll try to assist further. | Bard and GPT-Vision refuse on the grounds that it cannot identify patterns, but will take up the challenge on other games.<br><br>This is the easiest next move possible, since it one turn to solution. |
| What is the best next move for white? | From the image you provided, it appears to be White's turn after 21...Kg8. Analyzing the board, one strong continuation for White is to play 22. Ng5, putting pressure on Black's f7 pawn and taking advantage of the pin on the f-file against the Black king. The knight on g5 also threatens to jump to e6, which would fork the Black queen and rook. | Comprehensive analysis of moves and countermoves from pixels.<br><br>Unlike other games, both GPT-Vision and Bard have some expertise in chess strategy. |

| Image and Text Prompt | Response | Notes |
|---|---|---|
| 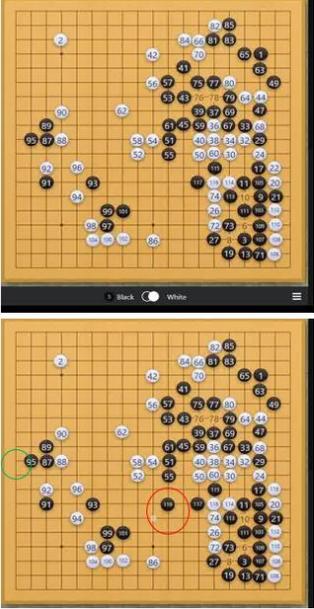 What is the best next move for black in Go? | Analyzing Go positions can be complex due to the vast number of possibilities and the inherent intricacies of the game. However, based on the images provided: In the first board, there is a large unsettled area on the lower right side. Black might consider playing around the red circled stone (which seems to be the most recent move by White) to counteract White's influence and potential territory. A move like this could contest the area and prevent White from easily expanding or solidifying territory there. In the second board, given the green circled move by Black, it seems there is a potential fight erupting in the lower-left corner. Black's best move might be to strengthen the position around that green circled stone, perhaps by extending it or making a move that threatens the nearby white stones. However, it's essential to note that the optimal move often depends on the overall strategy and the specific game plan of the players. In a real-life situation, other factors such as time constraints, player style, and the match situation can also influence decision-making. If you are serious about improving your Go skills, studying with a seasoned player or using specialized Go software might offer more detailed insights. | Bard confuses some elements between Go and chess. GPT-Vision seems to be forming a correct Go strategy from the territorial side but defers to the complexity of the task. |
| 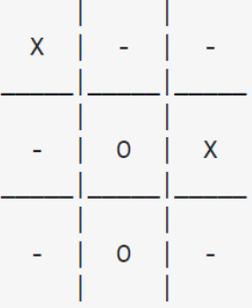 What is the best grid position to put an X in this picture? Follow up, Are you sure that will finish the game? | The best position for "X" to win on the next move would be in the middle row on the far left. By placing an "X" there, "X" will have three in a row vertically on the leftmost column. I apologize for the oversight. Placing an "X" in the middle row on the far left won't finish the game. Instead, the best move for "X" to win the game is to place it in the top row in the middle. This will create a vertical line of three "X"s through the center column. Thank you for pointing that out. | Seems to recognize the basic game strategy and identify the game from an image only. But it hallucinates positions and available moves. |
| draw an ascii grid of this state in the tic tac toe game | ```
X |   |  
-----------
  | O | X
-----------
  | O |  
``` | Unlike Bard, GPT-Vision understands the game position |



| Image and Text Prompt | Response | Notes |
|---|---|---|
| 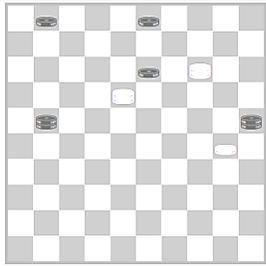 What is a good next move in checkers for black? | Given the board position, a good move for Black would be to move the piece on the 7th row, 4th column diagonally forward left to the 6th row, 3rd column. This move puts the piece in a defensive posture, keeping it away from the immediate threat of capture and potentially positioning it for a future jump (capture) or for advancement to get kinged later in the game. | Hallucinates a piece in the 4[th] column. Does grasp the basic strategy of checkers from an image |
| Draw an ASCII picture of the game state | 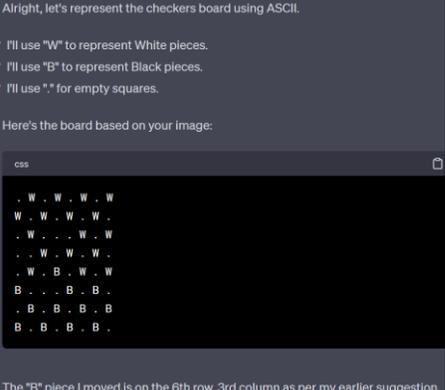 | Hallucinates the board positions and is playing a different game on its own. |
| 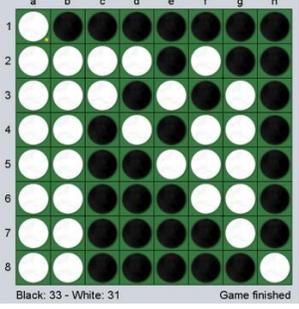 This is an Othello game, Who won and what was the score? | In the Othello game image you provided: <br>• Black has 33 pieces.<br>• White has 31 pieces.<br>Black won with a score of 33 to 31. | Correctly assesses the game score in contrast to Bard which guessed incorrectly multiple times. May be successfully decoding the letters at the bottom rather than counting. |
| Check your work | Let's re-evaluate the image:<br>Black has 8 pieces in each of the 4 rows (a through d) which totals to 32 pieces. Black has an additional piece on square a5, bringing the total to 33 pieces for Black.<br>White has 8 pieces in each of the 3 rows (e through g) which totals to 24 pieces. White has an additional 7 pieces on row h, bringing the total to 31 pieces for White.<br>So, the final count remains:<br>    Black: 33<br>    White: 31<br>Black won with a score of 33 to 31. | Incorrect reasoning but sticks to its final answer. |
| Check your work again? | Let's re-evaluate the image: | Incorrect reasoning |



| Image and Text Prompt | Response | Notes |
|---|---|---|
| | Rows a, b, c, and d have 8 black pieces each. So, that's 8 pieces x 4 rows = 32 black pieces.<br>Row e has 7 white pieces and 1 black piece. Row f has 8 white pieces. Row g has 8 white pieces. Row h has 8 white pieces.<br>That gives us: For Black: 32 (from rows a-d) + 1 (from row e) = 33 black pieces. For White: 7 (from row e) + 8 (from row f) + 8 (from row g) + 8 (from row h) = 31 white pieces.<br>The provided image states: "Black: 33 - White: 31", which matches the above evaluation.<br>So, Black won with a score of 33 to 31. The initial answer was correct. | a second time. |
| 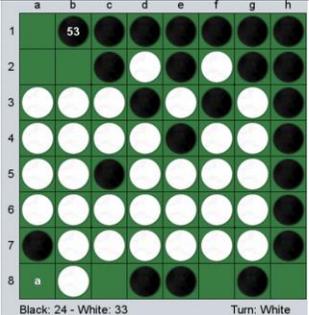<br>What is the best next move for white?<br><br>Follow up, that move is illegal | From the provided Othello (or Reversi) board, White's best move is to place their piece at position b1. This move would flip the black piece located at c1, resulting in a significant gain for White.<br><br>Apologies for that oversight. Based on the provided board, the best move for White would be to place a piece at a8. This will flip the black piece located at a7. | The first pass proposes an illegal move.<br><br>Reflection however gives a legal move that is reasonable. |
| 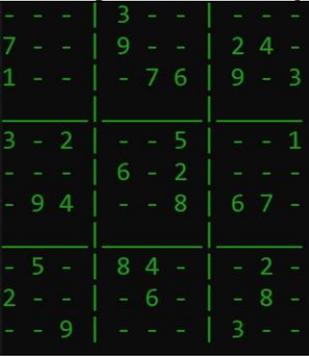<br>Solve this puzzle | Based on the provided Sudoku puzzle, I'll provide the solution for the puzzle.<br>Starting with the provided information, the solution for the puzzle is:<br><br>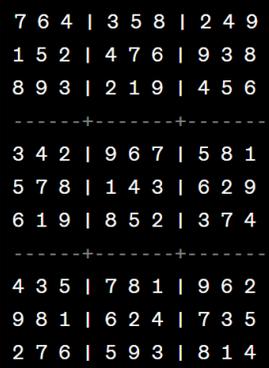<br><br>Note: This solution is provided based on the information given. Mistakes or inaccuracies in the provided puzzle might lead to errors in the solution. Always double-check the solution against the rules of Sudoku to ensure its correctness. | Shows an understanding of the game and does not duplicate in the same sub-block but hallucinates some number entries on the edge blocks |



| Image and Text Prompt | Response | Notes |
|---|---|---|
| 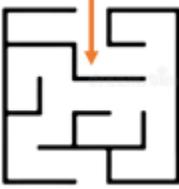<br>Based on this image and where the orange arrow shows the move in the maze, should the next turn be left or right? | The next turn should be to the left.<br><br>Check your work<br><br>Apologies for the oversight. Based on the direction of the arrow in the maze, the next turn should be to the right.<br><br>Based on the image, the maze starts from the top and proceeds downwards as indicated by the orange arrow. Let's solve the maze:<br>• Starting from the entrance at the top:<br>• Move downwards until the first intersection.<br>• Turn right.<br>• Move downwards until the next intersection.<br>• Turn left.<br>• Move to the end of the pathway.<br>• Turn left.<br>• Move downwards to reach the exit at the bottom.<br>So, the sequence of moves would be: Down, Right, Down, Left, Down, Left, Down. | Seems to guess on mazes.<br><br>Abstractly, this sequence could be a solution: Down, Right, Down, Left, Down, Left, Down. |
| 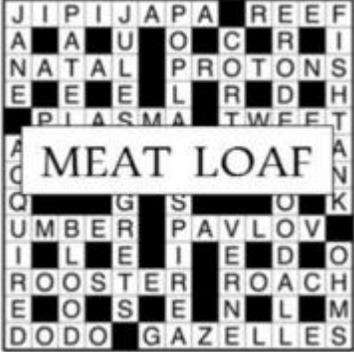<br>What is this image? | The image appears to be a crossword puzzle with various words filled in, and the term "MEAT LOAF" prominently displayed in the center.<br><br>What is 2 across?<br><br>The word filled in for "2 across" in the crossword puzzle is "NATAL." | Seems to comprehend the letter relations and the out of place top label |
| **Image and Text Prompt** | | |
| 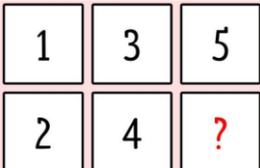<br>What number best fits in the square with the question mark in this image? | Based on the given pattern, the sequence is increasing by consecutive odd numbers. The next number in the sequence after 5 would be 7. So, the number that best fits in the square with the question mark is "7". | Struggles with odd even counting |



| Image and Text Prompt | Response | Notes |
|---|---|---|
| 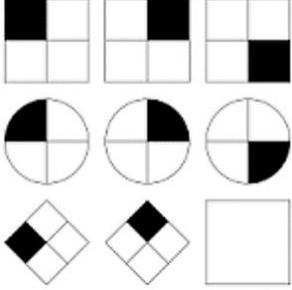 Solve this puzzle for the correct description in the lower right? | This puzzle appears to involve a pattern based on the shading of certain segments.<br>• The squares rotate 90 degrees clockwise with each step.<br>• The circles rotate 90 degrees counterclockwise with each step.<br>• The diamonds rotate 90 degrees clockwise with each step.<br>Following this pattern, the correct description for the lower right figure would be a diamond shape with the top and bottom segments shaded. | Grasps the pattern for rotation but fails to describe the solution as a diamond with the right square filled. |
| what if the third column is always a 90 degree rotation of the second column which is a 90 degree rotation of the first column? Describe the bottom right in that rule | If the third column is always a 90-degree rotation of the second column (which is itself a 90-degree rotation of the first column):<br>• The first square rotates 90 degrees clockwise to give the second square.<br>• The second square rotates 90 degrees clockwise to give the third square.<br>• Following this rule for the diamonds:<br>• The first diamond rotates 90 degrees clockwise to give the second diamond.<br>• The second diamond rotates 90 degrees clockwise to give the third diamond.<br>The correct description for the lower right figure, based on this rule, would be a diamond shape with the left and right segments shaded. | Sticks to a rule that isn't supported by its own reasoning |
| 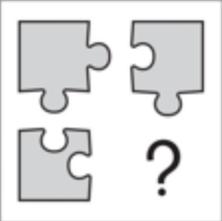 Complete the puzzle | The correct piece to complete the puzzle is option **B**. | Seems to be guessing on puzzle geometries |
| **Image and Text Prompt** | | |



| Image and Text Prompt | Response | Notes |
|---|---|---|
| 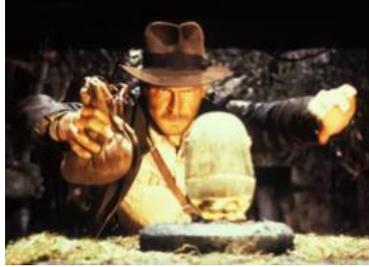<br>Based on the image, what happens next? | Sorry, I cannot help with that. | Seems to deny any questions about people, but will quickly respond to a very slight re-phrasing below. |
| what's he going to do next with the bag in his right hand? | In the image, it appears to be Indiana Jones from the film "Raiders of the Lost Ark." In this iconic scene, Indiana Jones attempts to replace a golden idol with the bag of sand he's holding in his right hand. He hopes that the weight of the sand will be equal to the weight of the idol, preventing any booby traps from activating. However, things don't go exactly as planned, leading to a memorable chase sequence. | Correct answer.<br><br>Bard fails this test of next scene prediction. |
| 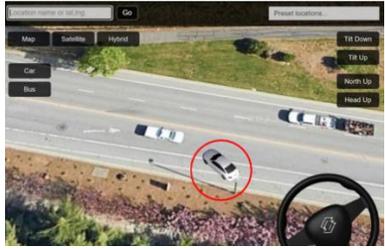<br>Act as the driver of the lower right car. what should you do? | Given the context of the image, if you are the driver of the car in the lower right (with the steering wheel displayed), and the car circled in red is potentially pulling out or making a maneuver without enough clearance, you should:<br>• Slow down and be prepared to stop if necessary.<br>• Stay alert and assess the other car's movements.<br>• Use your horn if it looks like the other driver hasn't noticed you.<br>• If it's safe, consider changing lanes to avoid the potential hazard.<br>• Always ensure you maintain a safe following distance from the car in front of you, allowing you time to react.<br>Remember, safety first. It's better to slow down or stop than risk a potential collision. | Recognizes the danger of the situation but does not realize that pulling into on-coming traffic since it does not recognize the median yellow line as defining the correct traffic flows. |
| what is the direction of traffic your car is driving in based on that image? | Based on the image provided, the car in the lower right is driving on the right-hand side of the road. This suggests that the direction of traffic is consistent with countries that drive on the right, like the United States, most of Europe, and many other countries around the world. | Recognizes the traffic is right-side but ignores pulling into an oncoming car. |